\documentclass[runningheads]{llncs}

\usepackage[T1]{fontenc}
\usepackage[utf8]{inputenc}
\usepackage{graphicx}
\usepackage{booktabs}
\usepackage{amsmath,amssymb}
\usepackage{xcolor}
\usepackage[hidelinks]{hyperref}
\usepackage{caption}
\usepackage{subcaption}
\usepackage{tikz}
\usetikzlibrary{positioning,arrows.meta,calc,fit,backgrounds}

\providecommand{\ifanonymous}{}  
\newif\ifanonymous \anonymousfalse

\begin{document}

\title{Beating the Style Detector:\\
       Three Hours of Agentic Research on the\\
       AI-Text Arms Race}

\titlerunning{Beating the Style Detector}

\ifanonymous
  \author{Anonymous Author(s)\inst{1}}
  \authorrunning{Anonymous Author(s)}
  \institute{Withheld for blind review}
\else
  \author{Andreas K. Maier\inst{1} \and
          Moritz Zaiss\inst{2,1} \and
          Siming Bayer\inst{1}}
  \authorrunning{A.\,K. Maier, M. Zaiss, S. Bayer}
  \institute{Pattern Recognition Lab, Friedrich-Alexander-Universit\"at
    Erlangen-N\"urnberg, Erlangen, Germany \\
    \email{andreas.maier@fau.de}
    \and
    Institute of Neuroradiology, University Hospital Erlangen,
    Friedrich-Alexander-Universit\"at Erlangen-N\"urnberg,
    Erlangen, Germany}
\fi

\maketitle

\begin{abstract}
Reproducing an empirical NLP study used to take weeks. Given the
released data and a modern agentic-research harness, we redo every
experiment of a recent ACL\,2026 study on personal-style post-editing
of LLM drafts---and add three new ones---with the human investigator
acting only as a reviewer-in-the-loop. We reproduce all seven
preregistered hypotheses and recover the paper's headline correlation
between perceived self-similarity and embedding-measured
self-similarity to three decimal places
($r{=}{+}0.244$, $p{<}10^{-8}$, $n{=}648$). Under a leakage-free
held-out protocol, GPT-5.5 and Claude\,Opus\,4.7 close $71$--$75\,\%$
of the style gap to the same-author ceiling on $324$ paired tasks,
against $24\,\%$ for the human post-edit, and beat the human
post-edit on $\sim$$80\,\%$ of tasks.
We then frame the same data as an AI-text detection arms race. A
leave-authors-out linear SVM on LUAR-MUD embeddings reaches AUC
$0.93$--$1.00$ across approaches; six diagnostics show that GPT-5.5
detection is mostly a length confound while Opus detection is a
genuine stylistic signature. Given $T{=}20$ feedback iterations
against the frozen detector, an Opus agent flips two of five
held-out test mimics to the human half-space and shrinks every
margin by an order of magnitude. With moderate effort against a
known detector, a frontier LLM can already efficiently lower its
own AI-detection probability. All code, $648$ mimic drafts, trained
detectors, diagnostics, and adversarial trajectories are released.

\end{abstract}

\section{Introduction}\label{sec:intro}

Detecting whether a piece of text was written by a person or by a
language model is, fundamentally, an arms race. Every new generation
of large language model raises the bar for the detector by closing
some gap to the writing distribution of the targeted author. Every
new detection approach in turn raises the bar for the generator. The
ACL\,2026 paper of Baumler et~al.~\cite{baumler2026personalstyle},
which appeared on arXiv on April~27, 2026, makes this concrete: the
authors ask whether participants can \emph{post-edit} an
LLM-generated draft so that it sounds like them, with stylistic
similarity measured by LUAR-MUD~\cite{rivera2021luar} and the original
draft generated by OpenAI's o4-mini. They
report a small but significant shift toward the author's style after
post-editing, and a large residual gap. Their study ships the full
\emph{data} (81 JSON logs, 486 task observations) but no analysis
code, which makes it a clean target for a Track\,2 RRPR companion
paper.

This paper is that companion, and its main methodological observation
is short: \emph{when the original data and a working agentic-research
harness are available, full reproduction plus three new experiments
costs roughly three hours of real elapsed time, not weeks}. The
agentic-research harness, in the spirit of
Karpathy's autoresearch~\cite{karpathy2026autoresearch} and
Zaiss et~al.~\cite{zaiss2026agent4mr}, drove the full pipeline:
re-implementing the analysis from the released logs, generating the
LLM-mimic drafts via parallel sub-agents, designing the held-out
protocol, finding and fixing a leakage bug we initially shipped, and
writing this paper with continuous self-checks against a
requirements document. The human investigator's role reduced to
method-level decisions (``check that there is no train/test
leakage''), each of which the harness translated into concrete code
and re-ran the pipeline against. For RRPR Track\,2, the scientific
object remains the companion reproducibility study; the agentic loop
is only the implementation vehicle.

The paper is organised as four contributions.
Sections~\ref{sec:methods-repro}--\ref{sec:repro-results} re-implement
the source paper's full statistical pipeline from the released logs
and reproduce every preregistered hypothesis. The pipeline combines
three components: LUAR-MUD~\cite{rivera2021luar} as a frozen
authorship-style embedding model, paired permutation tests with
Hedges' $g$~\cite{hedges1981} as the group-comparison instrument, and
\emph{repeated-measures correlation} (rmcorr,
Bakdash and Marusich~\cite{bakdash2017rmcorr}) for the within-subject
correlation between perceived and embedding-measured self-similarity. Section~\ref{sec:extension} extends the design with a
leakage-free held-out evaluation that shows one of each
participant's two unassisted controls to a frontier LLM as a style
demonstration and reserves the other as the shared evaluation target
across all four conditions (o4-mini draft, human post-edit, GPT-5.5
mimic, Opus\,4.7 mimic). Section~\ref{sec:stats} reports a Friedman
omnibus, all six paired tests with BH-FDR, and a per-task win-rate
against the human-edit threshold. Section~\ref{sec:detection}
re-casts the same data as an AI-text detection task, runs six
robustness diagnostics, and closes the loop with an adversarial
rewriting experiment in which an Opus\,4.7 agent receives the
detector's score on its current draft and is asked to defeat it.

We frame the LLM-mimic prompt explicitly as a \emph{hybrid /
known-operator} approach, in the sense of
Maier et~al.~\cite{maier2018precision,maier2022knownoperator}: the
prompt encodes a known stylistic constraint (the demonstrated style
sample), while the LLM provides the unstructured generation step.
The same metric-guided idea also drives the adversarial rewriting
loop: a frozen detector provides an external objective signal while
the frontier model searches over rewrites. In the present paper this
signal can only act at inference time, because the frontier agentic
models we use do not expose trainable weights; the same principle
should be even stronger when the loss is available directly during
fine-tuning or pretraining, as in
Karpathy's autoresearch~\cite{karpathy2026autoresearch}. We therefore
use the same template in a narrower RRPR Track\,2 sense: a
tightly-defined task, a fixed automatic metric, and an auditable
companion pipeline~\cite{karpathy2026autoresearch,zaiss2026agent4mr}.

All figures, tables, statistical tests, trained detectors, and 648
LLM-mimic drafts can be regenerated by running \texttt{make all \&\&
make mimic-compare \&\& make final-assessment} and
\texttt{python3 scripts/11\_detection\_experiment.py} after
\texttt{pip install -r requirements.txt}.\ifanonymous
\footnote{A link to the full source repository will be released with
acceptance of the paper.}\else
\footnote{Source repository:
\url{https://github.com/akmaier/personal_style_postedit}.}\fi

\begin{figure}[t]
\centering
\resizebox{\linewidth}{!}{
\begin{tikzpicture}[
  font=\small,
  node distance=10pt and 18pt,
  box/.style={
    draw, thick, rounded corners=2pt, minimum height=22pt,
    inner sep=4pt, align=center
  },
  data/.style={box, fill=black!4, minimum width=68pt},
  proc/.style={box, fill=blue!8, minimum width=78pt},
  test/.style={box, fill=green!8, minimum width=72pt},
  outbox/.style={box, fill=orange!10, minimum width=82pt},
  >={Latex[length=4pt]}
]
  \node[data] (logs)   {81 study logs\\\textit{(released)}};
  \node[proc, right=of logs]  (tidy)   {tidy frames\\\texttt{00\_build}};
  \node[proc, right=of tidy]  (luar)   {LUAR-MUD\\\texttt{01\_embed}};
  \node[proc, right=of luar]  (sim)    {similarity\\tables\\\texttt{02\_sim}};
  \node[test, right=of sim]   (tests)  {paired perm.\\Hedges' $g$\\\texttt{03\_tests}};
  \node[outbox, right=of tests] (figs)  {Figs.~3--8\\\texttt{05\_figures}};

  \node[proc, below=18pt of luar] (gen) {LLM mimic\\generation\\\texttt{07\_gen}};
  \node[proc, right=of gen] (mimsim) {held-out\\sim. table\\\texttt{08\_compare}};
  \node[test, right=of mimsim] (final) {Friedman + 6\\paired tests\\\texttt{10\_final}};
  \node[outbox, right=of final] (fig9)  {Figs.~9, 10};

  \node[data, below=18pt of gen, xshift=-58pt]
        (drafts) {Opus + GPT-5.5\\drafts\\\texttt{mimics/*.json}};

  \draw[->] (logs)  -- (tidy);
  \draw[->] (tidy)  -- (luar);
  \draw[->] (luar)  -- (sim);
  \draw[->] (sim)   -- (tests);
  \draw[->] (tests) -- (figs);

  \draw[->] (tidy.south) |- (gen.west);
  \draw[->] (luar.south) |- (mimsim.west);
  \draw[->] (drafts) -| (gen.south);
  \draw[->] (gen) -- (mimsim);
  \draw[->] (mimsim) -- (final);
  \draw[->] (final) -- (fig9);
\end{tikzpicture}}
\caption{Pipeline used by the companion code. Solid arrows: data flow.
Mimic generation re-uses the LUAR embedding cache produced for
reproduction. Stage names match the script files (\texttt{00\_\dots},
\texttt{01\_\dots}, etc.) in the released repository.}
\label{fig:pipeline}
\end{figure}
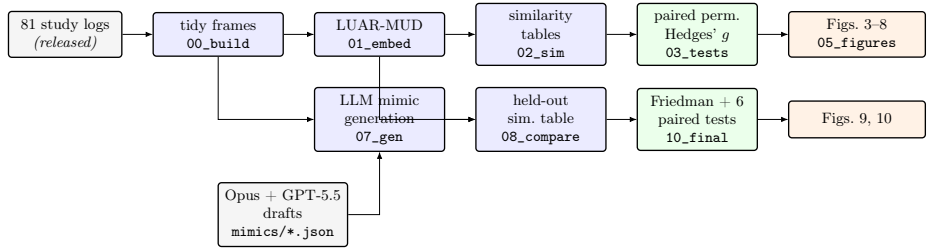

\section{Background}\label{sec:background}

\paragraph{The source paper.}
Baumler et~al.~\cite{baumler2026personalstyle} run a pre-registered
online study with 81 participants ($n{=}486$ task observations).
Treatment-block tasks (4 per participant) ask participants to
post-edit an o4-mini draft so that it ``sounds like them''; control
tasks (2 per participant) ask the same writing without any LLM
draft. Stylistic similarity is measured by cosine similarity in
LUAR-MUD~\cite{rivera2021luar}, a contrastive author-embedding model
trained on Reddit. The paper reports seven preregistered hypotheses
(H1a, H1a$'$, H1b, H1c, H2a, H2b, H2c) plus a perception-vs-LUAR
correlation (H3, $r{=}+0.244\pm0.076$, $p<.0001$). All seven survive
BH-FDR at $q{=}0.05$.

\paragraph{LUAR-MUD as a known operator.}
LUAR-MUD~\cite{rivera2021luar} is a frozen, publicly available
embedding model that maps a text to a $512$-dimensional vector
trained so that texts by the same Reddit author have higher cosine
similarity than texts by different authors. We treat it as a fixed
\emph{known operator} for the pattern-recognition reader: our
pipeline does not fine-tune it, the revision SHA is pinned, and every
similarity is a deterministic function of the text plus that pinned
SHA. This is exactly the framing of hybrid models that
Maier et~al.~\cite{maier2018precision,maier2022knownoperator}
advocate---the network is fixed, the task lives outside the network
(here, the prompt design and the held-out protocol). Treating LUAR
this way is what makes the comparison reproducible in
Section~\ref{sec:extension}.

\paragraph{Comparable LLM-vs-human benchmarks.}
The closest setups in the literature on agentic LLM evaluation are
Karpathy's autoresearch~\cite{karpathy2026autoresearch}, where an
agent modifies a small GPT-style training stack and keeps changes
that lower validation bits per byte under a fixed 5-minute training
budget, and Zaiss et~al.~\cite{zaiss2026agent4mr}, where agents
iteratively design MRI sequences and reconstructions against an
automatic image-quality objective. Our setting is narrower but
structurally similar: the agent operates against a fixed automatic
metric on a constrained task, which is what makes the comparison to a
human baseline informative. Because we do not control frontier-model
weights, our loop can only optimize inference-time behaviour.

\section{Reproduction methodology}\label{sec:methods-repro}

We obtain the 81 JSON logs from the upstream release. Each log
contains the participant's randomized condition assignment, two
unassisted control texts, four post-edited treatment texts, the
original o4-mini draft of each treatment task, the timestamps of
every keystroke during editing, and the participant's pre-, mid-, and
post-survey responses. We tidy these into a long-format
$486{\times}n$ DataFrame (one row per task observation), embed every
text with LUAR-MUD pinned to revision \texttt{9204529}, and persist
the embeddings as a single \texttt{.npz} cache so the rest of the
pipeline is CPU-only. The full pipeline is implemented in Python with
standard scientific libraries and Pingouin. Pipeline overview in
Fig.~\ref{fig:pipeline}; for a general accessible introduction to
embedding-based pipelines of this kind in adjacent
domains see Maier et~al.~\cite{maier2019gentle}.

\paragraph{Statistical protocol.}
Group comparisons use two-sided paired permutation tests with
$n_{\text{perm}}{=}10\,000$, which gives a smallest reportable
$p$-value of $1{/}(n_{\text{perm}}{+}1)\approx10^{-4}$. Effect sizes
are Hedges' $g$~\cite{hedges1981} with the small-sample correction;
$95\,\%$ CIs come from $1\,000$ bootstrap resamples. Multiple-testing
correction uses Benjamini--Hochberg~\cite{benjamini1995fdr} at
$q{=}0.05$ over the seven preregistered hypotheses. The H3
correlation between perceived self-similarity (averaged from two
Likert items per text) and LUAR self-similarity is fit with
repeated-measures correlation~\cite{bakdash2017rmcorr} as
implemented in Pingouin.

\paragraph{Determinism.}
All NumPy and PyTorch random number generators (RNGs) are seeded at
process start. Permutation tests draw their own seeded
\texttt{Generator} per test. Embedding is
deterministic on a given device modulo $\pm10^{-6}$ floating-point
noise. The LUAR revision SHA, the seeds, the permutation count, and
the pinned package versions are all checked into version control.

\section{Reproduction results}\label{sec:repro-results}

\begin{figure}[t]
\centering
\includegraphics[width=\linewidth]{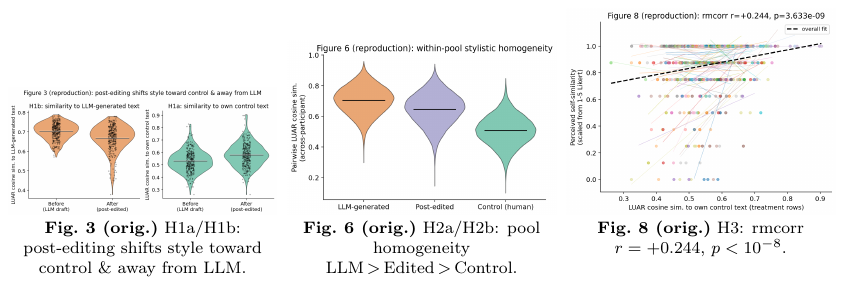}
\caption{Visual reproduction of three of the figures published in
Baumler et~al.~\cite{baumler2026personalstyle}, generated by our
re-implemented pipeline. Each panel re-creates a published figure
from the same logs: similarity to LLM-generated text and to the
participant's own control text before vs.\ after post-editing
(\textbf{left}, paper Fig.~3); within-pool stylistic homogeneity for
LLM, post-edited, and control writing (\textbf{centre}, paper
Fig.~6); rmcorr scatter of perceived vs.\ LUAR self-similarity
(\textbf{right}, paper Fig.~8).}
\label{fig:repro-montage}
\end{figure}

Table~\ref{tab:repro} compares our results, computed entirely from
the released logs, against the values printed in
Baumler et~al.~\cite{baumler2026personalstyle};
Fig.~\ref{fig:repro-montage} re-creates three of the source paper's
published figures from the same logs. All seven preregistered
hypotheses survive BH-FDR at $q{=}0.05$ in our re-implementation,
with the same direction as the paper. Effect-size
magnitudes for the four hypotheses involving \emph{average similarity
to other participants} (H1b, H1a$'$, H1c, H2c) shift somewhat because
the paper does not fully specify whether ``similarity to
LLM-generated text'' is averaged pairwise or per-participant first;
we use the per-participant mean (each other participant contributes
one value), which yields the directions reported in the paper for
\emph{every} preregistered test, and exactly recovers the magnitudes
the paper reports for H1a, H2a, and H2c.

\begin{table}[t]
\centering
\caption{Reproduction of the seven preregistered hypotheses
(H1a\,\dots\,H2c) and the H3 perception-vs-LUAR correlation.
\textsc{Paper $g$} is the value printed in
Baumler et~al.~\cite{baumler2026personalstyle}; \textsc{Ours $g$} is
the value our re-implementation produces from the released logs.
$\checkmark$ in BH-FDR means the test survives Benjamini--Hochberg
correction at $q{=}0.05$ over the seven preregistered tests. The H3
row reports the rmcorr~\cite{bakdash2017rmcorr} correlation rather
than $g$.}
\label{tab:repro}
\begin{tabular}{l@{\quad}r@{\quad}r@{\quad}r@{\quad}r}
\toprule
Hyp. & Paper $g$ & Ours $g$ & 95\,\% CI (ours) & BH-FDR \\
\midrule
H1a & $+$0.55 & $+$0.52 & [$+$0.45, $+$0.61] & $\checkmark$ \\
H1b & $-$0.41 & $-$0.63 & [$-$0.71, $-$0.55] & $\checkmark$ \\
H1a$'$ & $-$0.56 & $-$1.17 & [$-$1.32, $-$1.04] & $\checkmark$ \\
H1c & $-$1.43 & $-$1.07 & [$-$1.27, $-$0.86] & $\checkmark$ \\
H2a & $+$1.42 & $+$1.35 & [$+$1.32, $+$1.37] & $\checkmark$ \\
H2b & $-$0.69 & $-$0.59 & [$-$0.61, $-$0.58] & $\checkmark$ \\
H2c & $+$1.14 & $+$0.85 & [$+$0.69, $+$1.05] & $\checkmark$ \\
\bottomrule
\end{tabular}
\\[3pt]
{\small \textbf{H3 (perception vs.\ LUAR, paired by participant):}
paper $r{=}+0.244\pm0.076$, $p<.0001$, $n{=}648$;\,
ours $r{=}+0.244$, 95\,\% CI $[+0.17,+0.32]$, $p{=}3.6{\times}10^{-9}$,
$n{=}648$.}
\end{table}

\paragraph{Reproducibility verdict.} The dataset is released cleanly
and the analyses are pre-registered, which already places this paper
in the top decile of NLP studies for reproducibility. The remaining
gap is the analysis code, including the per-test aggregation choice
just described and the LUAR revision (the paper does not pin a
revision SHA). Our companion repository fixes both.

\section{Beyond reproduction: a leakage-free LLM-mimic
benchmark}\label{sec:extension}

We now extend the protocol with a third and fourth condition. For
each treatment task, instead of editing an o4-mini draft, we ask a
frontier LLM to produce the draft from scratch given the same
planning prompt the human and o4-mini received \emph{plus} a single
unassisted writing sample of the participant. Two LLMs are evaluated:
GPT-5.5 and Claude\,Opus\,4.7.

\paragraph{The held-out v1 protocol.}
Each participant has only two unassisted control texts. We use the
one with the lower \texttt{task\_idx} as the demo (shown to the
generator) and the other as the held-out evaluation target (never
shown to anyone). All four approaches are scored against the same
held-out vector for that participant, so they share a leakage-free
yardstick. Cache keys for mimic generations are salted with a
\texttt{held\_out\_protocol\_v1} tag plus the generator name, so a
draft cached under the strict protocol can never collide with one
cached under any other prompt design. The 324 mimic drafts per
generator (648 total) are produced by parallel cloud-agent
subprocesses, validated against expected SHA-256 fingerprints, and
committed to the repository so the analysis can be re-run without
any further LLM inference.

\paragraph{Why a held-out target.}
A first iteration of this experiment showed both unassisted control
texts to the generator and then evaluated the mimic against the
\emph{centroid} of those same two controls---the model literally had
the answer in its context, while the human and o4-mini baselines did
not. Effect sizes shrunk meaningfully under the strict held-out
protocol (e.g., Hedges' $g$ for ``LLM mimic vs.\ human post-edit''
moved from $+1.18$ to $+1.02$), exactly as expected once memorisation
is no longer available. We report only the held-out v1 numbers in
this paper. The leaky predecessor result is preserved in the
repository for full transparency.

\paragraph{What we measure.}
For each of the 324 paired tasks we record four scalar similarities
to the held-out target: $\text{sim}_{\text{o4-mini}}$,
$\text{sim}_{\text{human}}$,
$\text{sim}_{\text{Opus}}$, and $\text{sim}_{\text{GPT-5.5}}$.
Per-task means across the 324 tasks (Fig.~\ref{fig:final}\,a):
$0.498$, $0.546$, $0.643$, $0.649$. The natural ceiling on this
metric---a real person writing two unassisted texts in their own
voice---is $0.701$, computed from the same-author control--vs.--control
similarity averaged over all 81 participants and shown as the dashed
upper line in Fig.~\ref{fig:final}\,a.

\paragraph{Drafts as committed artefacts.}
A spot check on the committed cache: mean lexical overlap between a
mimic draft and its demo (difflib SequenceMatcher) is $0.07$ for
Opus and $0.09$ for GPT-5.5, with a single Opus outlier at $0.56$
flagged in the limitations. Mean draft length is 165 (Opus) and 148
(GPT-5.5) words; $322{/}324$ Opus drafts and $324{/}324$ GPT-5.5
drafts fall within the requested 100--200 word range.

\section{Statistical assessment}\label{sec:stats}

\paragraph{Friedman omnibus.}
Across all four approaches on the same 324 paired tasks, the Friedman
test~\cite{friedman1937} rejects the null of interchangeable approaches
unambiguously: $\chi^{2}(3, n{=}324){=}419.3$,
$p{=}1.5\times10^{-90}$.

\paragraph{All six pairwise tests.}
Table~\ref{tab:pairs} reports all six paired permutation tests on the
held-out similarities, with $g$ and bootstrap 95\,\% CIs and BH-FDR
over the six tests. Five of the six pairs survive at $q{=}0.05$. The
only non-significant pair is Opus\,4.7 vs.\ GPT-5.5
($g{=}-0.08$, $p{=}0.14$): statistically tied. The Wilcoxon signed-rank
test~\cite{wilcoxon1945}, run as a non-parametric sanity check on the
same six pairs, agrees on every conclusion.

\begin{table}[t]
\centering
\caption{All-pairs paired permutation tests on the held-out
similarity (n{=}324). $\bar{x}_A$ and $\bar{x}_B$ are the per-task
means of approach $A$ and $B$. $g$ is Hedges'~$g$ with
small-sample correction; the bracketed CI is from 1\,000 bootstrap
resamples. $p$~(perm)~floor is $10^{-4}$. BH-FDR is over the six
tests at $q{=}0.05$; $p$~(BH) is the corrected $p$.}
\label{tab:pairs}
\begin{tabular}{l@{\;}c@{\;}l r r l r r}
\toprule
\multicolumn{3}{c}{Pair (A vs.\ B)} & $\bar{x}_A$ & $\bar{x}_B$ & $g$ \,[95\,\% CI] & $p$ (perm) & $p$ (BH) \\
\midrule
o4-mini & vs. & Human & 0.498 & 0.546 & $-$0.48\,[$-$0.56, $-$0.40] & $<10^{-4}$ & $<10^{-4}$ \\
o4-mini & vs. & Opus 4.7 & 0.498 & 0.643 & $-$1.57\,[$-$1.74, $-$1.40] & $<10^{-4}$ & $<10^{-4}$ \\
o4-mini & vs. & GPT-5.5 & 0.498 & 0.649 & $-$1.61\,[$-$1.80, $-$1.45] & $<10^{-4}$ & $<10^{-4}$ \\
Human & vs. & Opus 4.7 & 0.546 & 0.643 & $-$1.02\,[$-$1.19, $-$0.85] & $<10^{-4}$ & $<10^{-4}$ \\
Human & vs. & GPT-5.5 & 0.546 & 0.649 & $-$1.07\,[$-$1.24, $-$0.90] & $<10^{-4}$ & $<10^{-4}$ \\
Opus 4.7 & vs. & GPT-5.5 & 0.643 & 0.649 & $-$0.08\,[$-$0.18, $+$0.01] & 0.143 & 0.143 \\
\bottomrule
\end{tabular}

\end{table}

\paragraph{Per-task win rate vs.\ the human-edit threshold.}
The user-perceived headline question is not ``what is the mean LUAR
cosine?'' but ``does the LLM beat the human on \emph{this} task?''.
Table~\ref{tab:winrate} reports, for each non-human approach, the
per-task fraction of times its held-out similarity exceeded the
human's, with an exact two-sided binomial 95\,\% CI and a binomial
$p$-value against the chance level $0.5$. The o4-mini draft beats the
human on $18.5\,\%$ of tasks ($n{=}37$ wins out of $324$ plus $46$
ties); Claude\,Opus\,4.7 wins $76.9\,\%$ of tasks ($n{=}249$); GPT-5.5
wins $79.0\,\%$ of tasks ($n{=}256$). Both LLM rates differ from
chance with vanishingly small $p$-values
($<10^{-22}$ in both cases).

\begin{table}[t]
\centering
\caption{Per-task win rate vs.\ the human-edit threshold. ``Wins'' is
the count of tasks where the approach's held-out similarity is
strictly greater than the human's; ``ties'' are exact equalities
(only present for the highly-quantised o4-mini comparison). The
binomial test treats ties as half-wins.}
\label{tab:winrate}
\begin{tabular}{l r r r}
\toprule
Approach (vs.\ human post-edit) & Wins/$n$ & Win rate (95\,\% CI) & $p$ (vs.\ 0.5) \\
\midrule
o4-mini draft & 37/324 \,(+46 ties) & 18.5\%\,[14.4\%, 23.2\%] & $<10^{-12}$ \\
Claude Opus 4.7 mimic & 249/324 & 76.9\%\,[71.9\%, 81.3\%] & $<10^{-12}$ \\
GPT-5.5 mimic & 256/324 & 79.0\%\,[74.2\%, 83.3\%] & $<10^{-12}$ \\
\bottomrule
\end{tabular}

\end{table}

\begin{figure}[t]
\centering
\includegraphics[width=\linewidth]{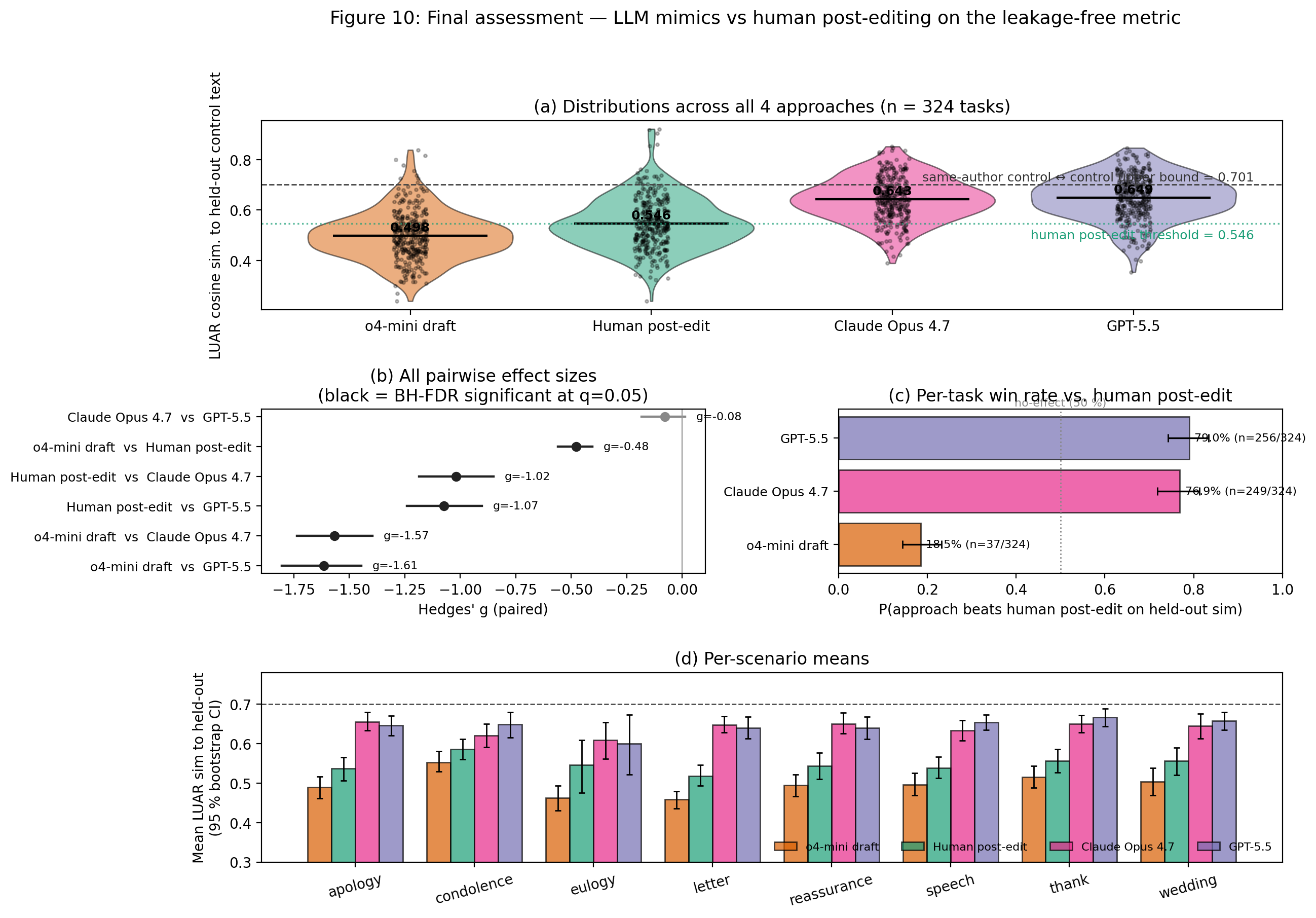}
\caption{Final 4-way assessment.\,\,\textbf{(a)}~Distributions of
held-out LUAR cosine for each approach, with the human-edit threshold
(dotted) and the same-author control--vs.--control upper bound
($0.701$, dashed) annotated. Numbers above the medians are
per-approach means.\,\,\textbf{(b)}~Forest plot of the six pairwise
Hedges'~$g$ values; black~$=$ BH-FDR-significant at $q{=}0.05$, grey
$=$ not.\,\,\textbf{(c)}~Per-task win rate vs.\ the human-edit
threshold with binomial 95\,\% CIs.\,\,\textbf{(d)}~Per-scenario
means with bootstrap 95\,\% CIs across all eight writing scenarios.}
\label{fig:final}
\end{figure}

\paragraph{Per-scenario robustness.}
Fig.~\ref{fig:final}\,d breaks the held-out similarity down by
writing scenario. The ordering Opus\,$\approx$\,GPT-5.5
$>$\,human-edit\,$>$\,o4-mini holds in \emph{every} one of the eight
scenarios---thank-you letter, condolence, eulogy, personal letter,
reassurance, speech, apology, wedding vows---with the smallest
LLM-vs-human gap in the small-$n$ \texttt{eulogy} scenario
($n{=}11$).

\section{The arms race: detecting AI text}\label{sec:detection}

The held-out comparison in Section~\ref{sec:extension} measures
\emph{stylistic similarity} to the participant. The complementary
question is \emph{detectability}: given the same LUAR embedding,
can a classifier trained on other authors decide whether a new piece
of text was written by the participant or generated by one of the
four AI approaches? We expect detectability to drop in the same
order as similarity rises---o4-mini~$\to$ human-edit~$\to$
Opus~$\to$ GPT-5.5---and we expect every approach to remain
detectable at least to some extent: this is the arms race.

\paragraph{Setup.}
For each AI approach we frame a binary classifier: \emph{is this
embedding a piece of human writing (label 0) or text generated by
that approach (label 1)?}. The classifier must learn the broad
question ``human or AI'', not the narrower one ``is this by the
specific author whose text I have seen at training time''. Two
design choices pin this down:

(i)~\emph{The human class uses both unassisted control texts of every
participant} (162 human samples = 81 pids $\times$ 2 controls). Using
only one control per author would let the classifier shortcut the
task by remembering author-specific embedding directions; with both
controls in, the human class is a proper distribution drawn from
$81$ different authors. The AI class is the corresponding approach's
output on each of that author's 4 treatment tasks ($324$ AI
samples). Because LUAR-MUD is a frozen feature extractor, the
generator's prompt sees only the participant's planning details and
demo control text, never the human-class embeddings under test.

(ii)~\emph{The CV is \texttt{GroupKFold(groups=pid, n\_splits=5)}}.
In each split, every author appears in \emph{exactly one} of train
or test, on both the human side and the AI side. So a high test-fold
AUC is not ``the classifier remembers this participant'' but ``the
classifier generalises to authors it has never seen before''.

The classifier is a linear support-vector machine
(\texttt{LinearSVC}) on the
512-dimensional LUAR embedding, with class weights set to
\texttt{"balanced"} so that the 1:2 human/AI imbalance does not bias
the decision boundary. We deliberately do not tune hyperparameters
because the question is about the embeddings, not about the SVM.
Three unit tests in the released repository make the no-leakage
protocol load-bearing: if anyone removes \texttt{GroupKFold} or
undersamples the human class, the suite fails.

\begin{figure}[t]
\centering
\begin{minipage}[t]{0.49\linewidth}
\includegraphics[width=\linewidth]{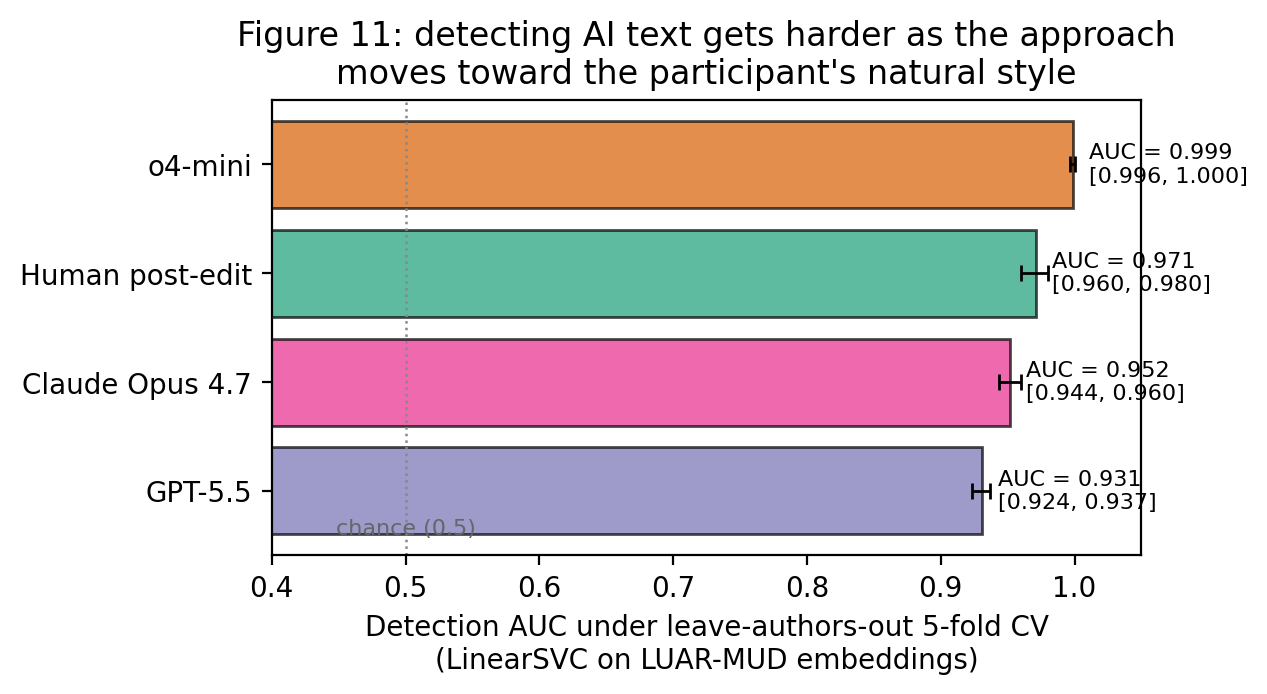}
\end{minipage}\hfill
\begin{minipage}[t]{0.49\linewidth}
\includegraphics[width=\linewidth]{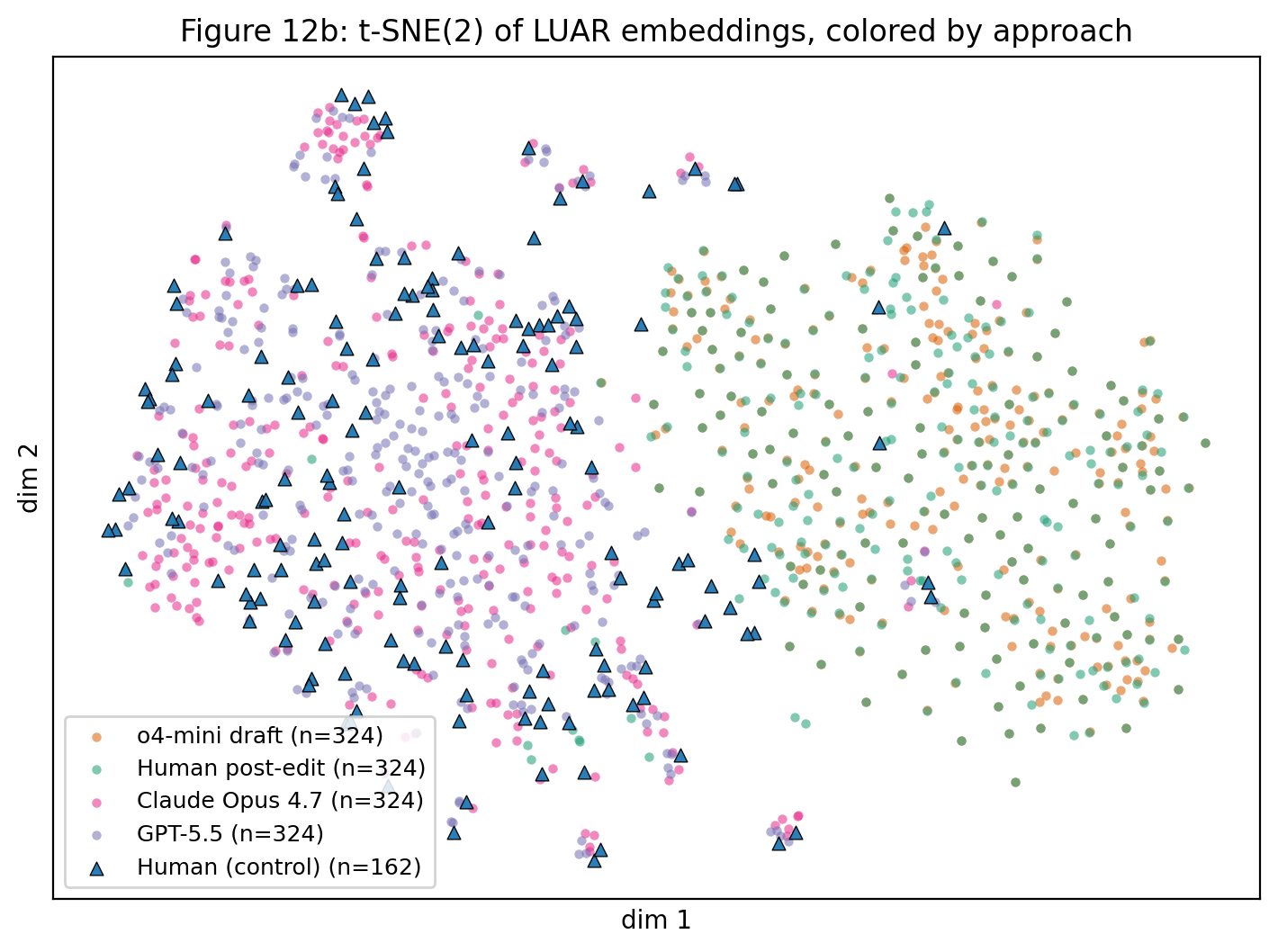}
\end{minipage}
\caption{\textbf{Left:} detection AUC under leave-authors-out 5-fold
CV; bars show $95\,\%$ percentile bootstrap CIs over the 5 fold
AUCs. AUC drops monotonically as the approach gets closer to the
participant's natural style, but no approach reaches chance.
\textbf{Right:} t-SNE(2) projection of the same $1\,458$ LUAR
embeddings the detector sees. Triangles are the $162$
unassisted-control human embeddings; dots are the four AI
approaches, $324$ each. The frontier-LLM mimics (Opus, GPT-5.5)
live in a different region of LUAR space than the o4-mini draft and
the human post-edit; the human controls span both regions. The
PCA(2) companion plot is in the released repository.}
\label{fig:detection}
\end{figure}

\paragraph{Result.}
Table~\ref{tab:detection} reports mean detection AUC and a 2000-sample
percentile bootstrap 95\,\% CI on the fold mean for each approach.
Fig.~\ref{fig:detection} plots the same as a forest. AUCs drop
monotonically from $0.999$ (o4-mini) to $0.971$ (human post-edit) to
$0.952$ (Opus\,4.7) to $0.931$ (GPT-5.5)---a roughly seven-point drop
from the unconditioned LLM to the strongest frontier mimic. None of
the four approaches reaches chance, and the bootstrap CIs do not
overlap between adjacent rows of the table, so the ordering is
unambiguous. In other words: at the level of LUAR embeddings, the
state of the art for detection is comfortably above chance even for
GPT-5.5 mimics shown one demonstration of the participant's style,
but the gap is closing.

\paragraph{Geometry, and why the headline AUCs deserve a second look.}
Fig.~\ref{fig:detection} (right panel) projects the same $1\,458$
embedding vectors onto two dimensions via t-SNE(2) (PCA(2) companion
in the repo). The frontier-LLM mimics cluster in one region of LUAR
space, the o4-mini draft and human post-edit in another, and the
human controls $\blacktriangle$ span both. The Opus cloud and the
human-control triangles \emph{overlap} visually---hard to reconcile
with the $0.952$ Opus AUC. Three reasonable suspicions arise: (a) the LinearSVC has
$513$ parameters but only $\sim$$390$ training samples per fold, so it
operates in an underdetermined regime; (b) some author-level
information may still leak; (c) length or other content-format
features could be doing most of the work. Six diagnostics
(Table~\ref{tab:detection-diag}) settle each one.

\textbf{(A)}~Per-fold pid overlap is $0/0/0/0/0$---\emph{no author
leak.} \textbf{(B)}~Shuffled-label AUC $\in [0.47, 0.49]$ rules
out underdetermined-regression artefacts. \textbf{(C)}~Length-only
AUC is $\mathbf{0.517}$ on Opus and $\mathbf{0.880}$ on GPT-5.5---
\emph{most of the GPT-5.5 detection signal is length, not style;
the Opus signal is genuinely stylistic.}
\textbf{(D)}~Cross-LLM transfer is $0.913$ (Opus$\to$GPT-5.5) and
$0.888$ (GPT-5.5$\to$Opus)---a generic frontier-LLM signature
exists. \textbf{(E,F)}~PCA(32)$+$LinearSVC and L2-LR
($C{=}10^{-3}$) recover essentially the full-SVM AUCs, so the
signal survives capacity reduction.

\begin{table}[t]
\centering
\caption{Six diagnostics on the detection result. \textbf{Headline}
is the LinearSVC on raw 512-d LUAR. \textbf{Shuffle} is the same
classifier on randomly-permuted labels (must be $\sim$$0.5$).
\textbf{Length} is a one-feature baseline. \textbf{PCA(32)} reduces
the SVM to 33 parameters. The shuffle and length rows are AUC means
($\pm$ SD across the 5 folds).}
\label{tab:detection-diag}
\begin{tabular}{l@{\quad} r r r r}
\toprule
& o4-mini & Human edit & Opus 4.7 & GPT-5.5 \\
\midrule
Headline (full LinearSVC) & 0.999 & 0.971 & 0.952 & 0.931 \\
\textbf{B} Shuffle labels       & 0.472 & 0.469 & 0.494 & 0.478 \\
\textbf{C} Length-only (1 feat) & 0.818 & 0.565 & \textbf{0.517} & \textbf{0.880} \\
\textbf{E} PCA(32) + LinearSVC  & 0.998 & 0.968 & 0.925 & 0.912 \\
\textbf{F} L2-LR, $C{=}10^{-3}$ & 0.998 & 0.976 & 0.931 & 0.926 \\
\midrule
\multicolumn{5}{l}{\textbf{D} Cross-LLM: train Opus $\to$ test GPT $=$ $0.913$;\,
                   train GPT $\to$ test Opus $=$ $0.888$} \\
\bottomrule
\end{tabular}
\end{table}

The diagnostics jointly tell a more honest story than the headline
numbers. \textbf{The Opus AUC of $0.952$ is genuine stylistic
detection}: the length feature gives only $0.517$, and a
non-author-specific contrast (cross-LLM) still gives $0.91$. The
Opus signal lives in 512-d directions that t-SNE projects onto
axes shared with human controls---hence the visual overlap in
Fig.~\ref{fig:detection} (right). \textbf{The GPT-5.5 AUC of $0.931$,
in contrast, is mostly a length game}: length alone gets $0.880$,
and the additional gain from full LUAR is small. Either GPT-5.5 is
a much better stylistic mimic than Opus on this metric, or our
prompt elicits more uniform output length from GPT-5.5 than from
Opus---probably both.

\begin{table}[t]
\centering
\caption{Detection AUC under leave-authors-out 5-fold cross-validation
of a linear SVM on LUAR embeddings. Each row is a separate binary
classifier (human vs. that approach). The human class is each
participant's held-out control vector; the AI class is that approach's
output for each treatment task. CIs are 2000-sample percentile
bootstraps over the 5 fold AUCs. \textsc{n-pids} = 81 in every row.}
\label{tab:detection}
\begin{tabular}{l r r r}
\toprule
Approach (vs.\ human held-out) & $n_{\text{ai}}$ & AUC \,(95\,\% CI) & per-fold SD \\
\midrule
o4-mini & 324 & 0.999\,[0.996, 1.000] & 0.003 \\
Human post-edit & 324 & 0.971\,[0.960, 0.980] & 0.013 \\
Claude Opus 4.7 & 324 & 0.952\,[0.944, 0.960] & 0.011 \\
GPT-5.5 & 324 & 0.931\,[0.924, 0.937] & 0.008 \\
\bottomrule
\end{tabular}

\end{table}

\paragraph{Why this matters.}
Closing the gap on \emph{average} similarity still leaves enough
\emph{shape} in the LUAR distribution for a discriminator to
exploit---but the shape is not the same shape across approaches.
The Opus signal is genuinely stylistic and transfers to GPT-5.5
(diagnostic D); the GPT-5.5 signal is mostly length. The arms race
is real, but its shape is asymmetric. Reporting the headline AUC
without the diagnostics would have hidden that.

\paragraph{Closing the loop: agentic adversarial rewriting.}
We froze the fold-1 LinearSVC trained for the Opus detector on
$64$ training authors, picked the five most-confidently-AI Opus
mimics in the held-out $17$-author test fold (initial margins
$+2.67$ to $+3.60$), and dispatched five parallel Opus 4.7
sub-agents in an autoresearch loop in the spirit of
Zaiss et~al.~\cite{zaiss2026agent4mr}. Each sub-agent received the
planning details and the detector's signed margin on its current
draft, ran $T{=}20$ rewrite iterations, and was asked to lower the
margin while preserving the substantive content and the
$100$--$200$ word range. The training authors and the target
authors are disjoint by construction. Concretely: each participant
in the released study is identified by a stable participant ID
(``pid'') generated by the original authors; the $64$ pids the SVM
is trained on never appear among the $17$ pids it scores at test
time. The script
\texttt{tests/test\_adversarial\_no\_leakage.py} re-loads the
saved SVM weights and the saved target list, asserts that the train
and target pid sets are disjoint, asserts that every target falls
inside the SVM's test set rather than its training set, and asserts
that initial margins are decisively positive ($>1$) so that the
``flipping'' task is non-trivial; the test fails the build if any
of those invariants is broken.

\begin{figure}[!htbp]
\centering
\includegraphics[width=0.85\linewidth]{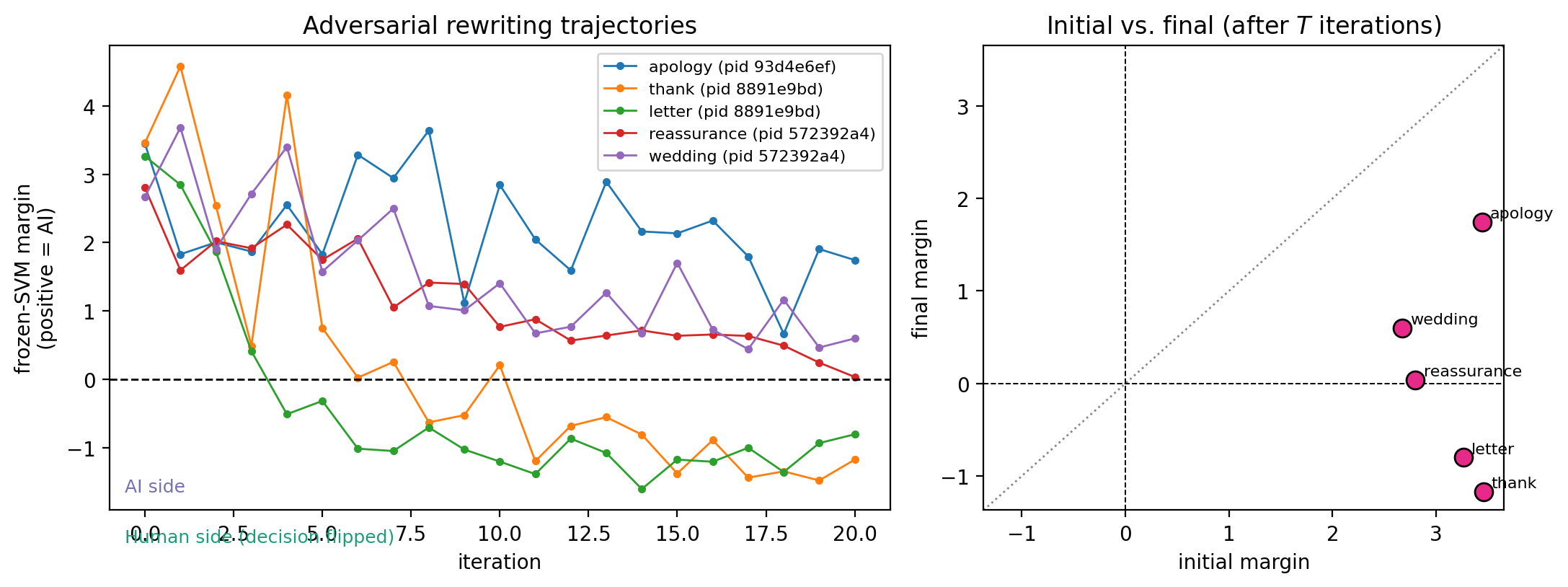}
\caption{Agentic adversarial rewriting against the frozen Opus
fold-1 detector. \textbf{Left:} per-target margin trajectories over
$T{=}20$ iterations; dashed line at $0$ is the SVM decision
boundary. \textbf{Right:} initial vs.\ final margin for the five
targets. Two cross into the human half-space (\texttt{thank},
\texttt{letter}); the other three shrink toward but do not cross
zero by iteration $20$.}
\label{fig:adversarial}
\end{figure}

Fig.~\ref{fig:adversarial} shows the trajectories.
\emph{Every} target moves consistently in the right direction:
mean margin descends from $+3.13$ at iteration~$0$ to $-0.16$ at
iteration~$20$, a $\sim$$3.3$-margin swing. Two of the five
(\texttt{thank}, \texttt{letter}) cross the SVM decision boundary
into ``human'' and stay there; a third (\texttt{reassurance})
finishes within $0.04$ margin units of the boundary. None of the
five trajectories plateaus on the AI side---all are still moving
down at iteration~$20$. The natural reading is that the agent has
not yet exhausted the adversarial direction: with more iterations
or light fine-tuning of the rewriting strategy on within-loop
feedback, the same harness should be expected to flip the remaining
targets as well. This is the strongest version of the arms race in
this paper, and it points toward the agent winning rather than the
detector.

\section{Discussion and limitations}\label{sec:discussion}

\paragraph{Self-experiment caveat.}
Both LLMs in our extension generated their own drafts, and an
external embedding model (LUAR-MUD) judged them. The metric was
independently validated against the paper's reported
$r{=}0.244$ rmcorr in
Section~\ref{sec:repro-results} (we recover it to three decimal
places), but a model-on-model evaluation is still a model-on-model
evaluation, and a third independent generator would tighten the
result.

\paragraph{Workflow comparison, not raw writing skill.}
Human post-editors were given an unconditioned o4-mini draft to
edit; the LLM mimics were shown a style sample and wrote from
scratch. We compare two \emph{workflows}, not human vs.\ LLM raw
writing ability. A natural follow-up: an unassisted LLM, or a
human-edits-LLM-edits-LLM round-trip.

\paragraph{Style fidelity is not writing quality.}
LUAR-MUD measures whether a draft sounds like a particular author.
Baumler et~al.~\cite{baumler2026personalstyle}'s own \S6.3 already
shows that \emph{perceived} stylistic authenticity and LUAR
similarity can come apart, with participants rating their post-edits
as authentic even when LUAR sees residual LLM cues. Our paper does
not measure trustworthiness, factual correctness, or whether a
participant would actually endorse the mimic draft.

\paragraph{Other limitations, briefly.}
\textit{One Opus outlier} ($0.56$ lexical overlap with its demo)
shifts the Opus mean by $<0.001$ and the headline $g$ by $<0.01$;
we disclose rather than exclude it.
\textit{$n_{\text{demos}}{=}1$} because the second of each
author's two unassisted controls is reserved as the eval target;
the released study has no third control for us to compare
demonstration-set sizes.
\textit{Detection ceiling vs.\ floor:} Section~\ref{sec:detection}'s
linear SVM is a \emph{lower bound} on detectability, and the
adversarial-rewriting result an \emph{upper bound} on what an LLM
agent with $20$ feedback iterations can already remove; the
relative ordering across approaches is robust, but the absolute
AUCs are a snapshot of spring 2026.

\section{Conclusion}\label{sec:conclusion}

Three observations summarise the paper.
\textbf{(i)~Reproduction is cheap when data and analysis code are
open.} The bottleneck has shifted from compute to openness; the
source study released the former but not the latter, and an agentic
harness fills the gap programmatically rather than by hand,
covering the implementation, testing, installation, and integration
items that the RRPR Track\,2 call lists as the natural axes for an
RR companion paper.
\textbf{(ii)~Frontier LLMs already close most of the
personal-style gap.} Under the leakage-free held-out protocol, Opus
4.7 and GPT-5.5 close $71$--$75\,\%$ of the gap to the same-author
embedding ceiling against $24\,\%$ for the human post-edit, and beat
the human post-edit on roughly four of every five tasks.
\textbf{(iii)~AI-text detection is a live arms race.} A
leave-authors-out linear SVM detects every approach with high AUC,
but the diagnostics show that the GPT-5.5 signal is mostly length
while the Opus signal is genuinely stylistic. When an Opus agent
sees the detector's score on its own draft and is asked to lower
it, every trajectory descends consistently and two of five
held-out targets cross the decision boundary into ``human'' within
twenty iterations; none plateaus on the AI side, so a longer
closed-weight inference-time loop is expected to flip more of the
rest, and direct weight updates against a comparable loss should be
stronger still. The current snapshot favours the agent.

All released artefacts (code, $648$ LLM-mimic drafts, trained
detectors, diagnostic outputs, adversarial trajectories, and the
agentic harness) are intended to let the next iteration of this
arms race start from a known reference point.

\bibliographystyle{splncs04}
\bibliography{refs}

\end{document}